\def\year{2021}\relax
\documentclass[letterpaper]{article} 
\usepackage{arxiv_sty}  
\usepackage{times}  
\usepackage{helvet} 
\usepackage{courier}  
\usepackage[hyphens]{url}  
\usepackage{graphicx} 
\usepackage{natbib}  
\usepackage{caption} 
\usepackage{bm,amsmath,amssymb}
\usepackage{xspace}
\usepackage{graphicx}
\usepackage{color}
\usepackage{algorithm}
\usepackage{algorithmic}
\usepackage{colortbl}
\definecolor{light_gray}{RGB}{170,170,170}
\usepackage[utf8]{inputenc} 
\usepackage[T1]{fontenc}    
\usepackage{url}            
\usepackage{booktabs}       
\usepackage{amsfonts, mathtools}       
\usepackage{nicefrac}       
\usepackage{microtype}      
\usepackage{longtable,lscape}
\usepackage{array,multirow}
\usepackage[bb=boondox,bbscaled=.95]{mathalfa}
\usepackage{subcaption}
\usepackage{wrapfig}
\usepackage{enumitem}
\usepackage{bm}
\usepackage{physics}

\newcommand{\cL}{\mathcal{L}}

\newcommand{\nH}{{n_h}}
\newcommand{\nX}{{n_x}}
\newcommand{\nA}{{n_a}}
\newcommand{\nZA}{{n_{z, a}}}
\newcommand{\nZX}{{n_{z, x}}}
\newcommand{\nTh}{{n_\theta}}
\newcommand{\nPh}{{n_\phi}}
\newcommand{\nPs}{{n_\psi}}
\newcommand{\nOm}{{n_\omega}}

\newcommand{\Ab}{\mathbf{A}}

\newcommand{\Xb}{\mathbf{X}}

\newcommand{\ab}{\mathbf{a}}

\newcommand{\hb}{\mathbf{h}}

\newcommand{\xb}{\mathbf{x}}

\newcommand{\zb}{\mathbf{z}}

\newcommand{\cT}{\mathcal{T}}
\newcommand{\R}{\mathbb{R}}

\definecolor{diffeq_blue}{RGB}{11,132,147}

\newcommand{\model}{IMODE\xspace}

\title{Neural Ordinary Differential Equations for Intervention Modeling}
\author {
    Daehoon Gwak, \hspace{-1.8mm} \textsuperscript{\rm 1}
    Gyuhyeon Sim, \hspace{-1.8mm} \textsuperscript{\rm 2}
    Michael Poli, \hspace{-1.8mm} \textsuperscript{\rm 2}
    Stefano Massaroli, \hspace{-1.8mm} \textsuperscript{\rm 3}
    Jaegul Choo, \hspace{-1.8mm} \textsuperscript{\rm 2}
    Edward Choi \hspace{-1.8mm} \textsuperscript{\rm 2}
}
\affiliations {
    \textsuperscript{\rm 1} Korea University,
    \textsuperscript{\rm 2} KAIST,
    \textsuperscript{\rm 3} The University of Tokyo \\
    hune282@korea.ac.kr, \{ghsim, poli\textunderscore m, jchoo, edwardchoi\}@kaist.ac.kr, massaroli@robot.t.u\text{-}tokyo.ac.jp
}
\begin{document}

\maketitle

\begin{abstract}
By interpreting the forward dynamics of the latent representation of neural networks as an ordinary differential equation, Neural Ordinary Differential Equation (Neural ODE) emerged as an effective framework for modeling a system dynamics in the continuous time domain.
However, real-world systems often involves external interventions that cause changes in the system dynamics
such as a patient being administered with particular drug.
Neural ODE and a number of its recent variants, however, are not suitable for modeling such interventions as they do not properly model the observations and the interventions separately.
In this paper, we propose a novel neural ODE-based approach (\texttt{\model}) that properly model the effect of external interventions by employing two ODE functions to separately handle the observations and the interventions.
Using both synthetic and real-world time-series datasets involving interventions, our experimental results consistently demonstrate the superiority of \model compared to existing approaches.
\end{abstract}

\begin{figure}
\centering
\includegraphics[width=0.4\textwidth]{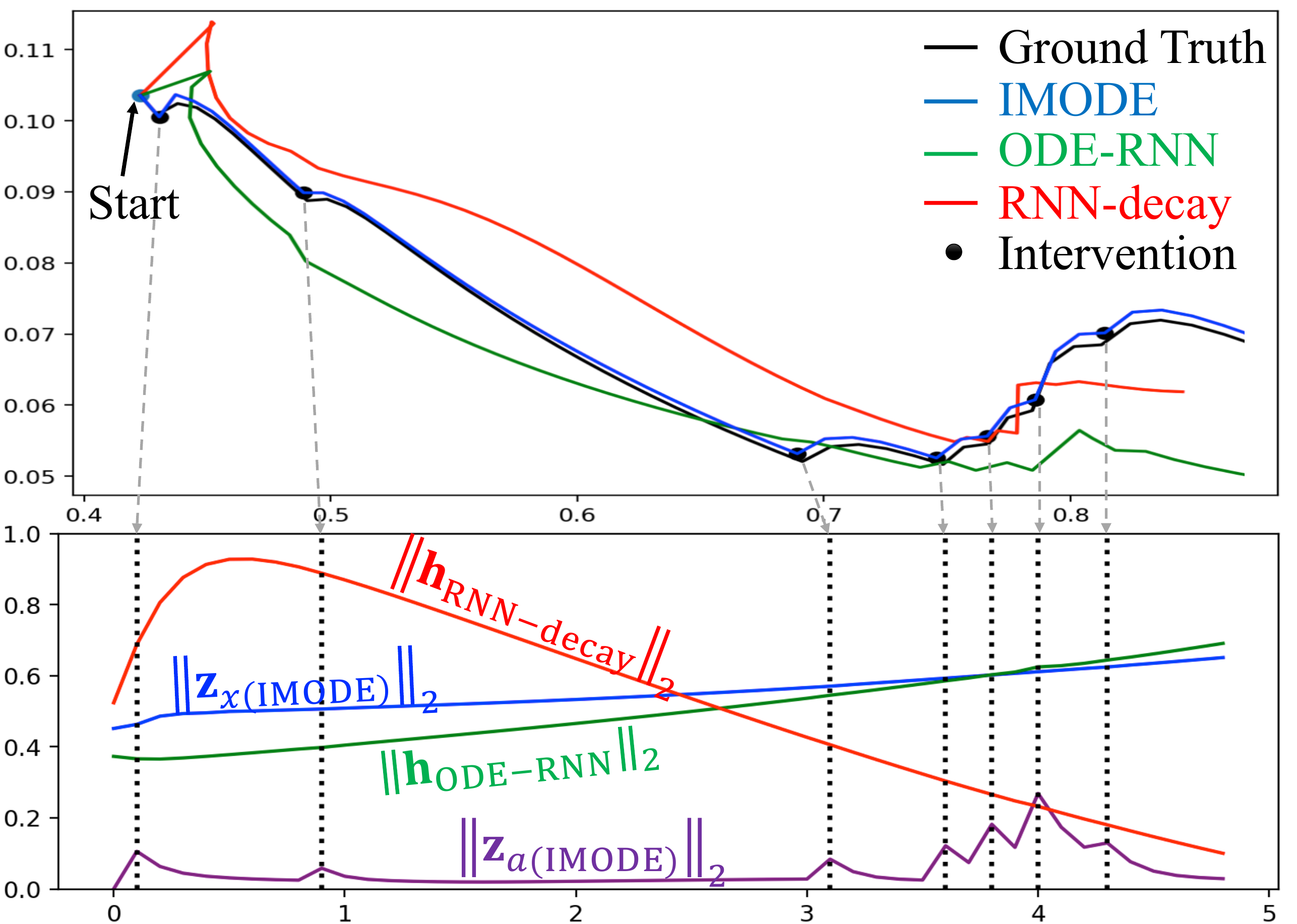}
\caption{\small(\textit{Top}) In a 2D plane, \model, ODE-RNN and RNN-Decay simulate a trajectory intervened by time-decaying effects.
(\textit{Bottom}) X-axis indicates time while Y-axis denotes $L_2$ norm of hidden-state vectors. \model separately models the autonomous dynamics ($||\zb_{x \text{(\model)}}||_2$) and the intervention effect ($||\zb_{a \text{(\model)}}||_2$) (see Figure~\ref{fig:architecture}), while ODE-RNN and RNN-Decay fail to handle the interventions correctly.\vspace{4mm}}
\label{fig:motivation}
\end{figure}

\section{Introduction}
\label{sec:intro}
Although we live in continuous time, physical systems (\textit{e.g.}, bouncing ball, patient state) are often observed in a discretized fashion, either regularly or irregularly.
For example, while climate sensors can collect information at every hour, patient blood samples are drawn only by a physician's order.
Various approaches have been proposed to handle such time-series data with neural networks, often modifying recurrent neural networks (RNNs) with varying degrees of success \cite{choi2016doctor, du2016recurrent, lipton2016directly, baytas2017patient, che2018recurrent}, until Neural Ordinary Differential Equations (Neural ODEs) \cite{chen2018neural} proposed a natural framework to model a system dynamics in a  continuous time domain.

Neural ODEs view the forward pass of the vector representation $\hb$, often corresponding to the system state, as numerically solving an ordinary differential equation using the time derivative $\dot \hb:={d\hb}/{dt}$ parameterized by a neural network.
As this framework provides a natural means to handle both regular as well as irregular time-series data, previous studies have extended Neural ODEs to encode observation sequences.
These approaches have demonstrated improved performance in prediction tasks on simulated data, climate records, and medical records \cite{rubanova2019latent, de2019gru, jia2019neural, poli2019graph}.

Real-world systems, however, are often influenced by external factors (\textit{i.e. interventions}), such as a patient being administered some medication at a particular time.
Depending on the system characteristic, these influences can change the system dynamics instantaneously or in a prolonged manner.
While effective in modeling intervention--free dynamics, previous approaches capable of handling discrete or continuous feature evolution such as RNN--decay, GRU-D, and ODE--RNNs \cite{che2018recurrent, rubanova2019latent} were not designed to handle cases with external interventions.
In particular, ODE--RNNs and their derivative architectures \cite{de2019gru, jia2019neural} place a strong assumption on the effect of additional observations on the system state;
by aggregating input information with either recurrent cells or a multi-layer preceptron (MLP), the system state is directly modified by each observations.

In response, this paper proposes Intervention-Modeling Ordinary Differential Equation (\texttt{\model}) which aims to model systems with (regular or irregular) interventions.
Unlike alternative approaches, \model is designed to handle interventions that affect the system dynamics in various ways.
Specifically, we employ two separate ODE functions, where one is tasked with learning the autonomous dynamics from a sequence of observations, and the other is tasked with learning the effects of external forces on the system.
Delegating the task of modeling the intervention effect to a separate component ultimately leads to a disentangled, interpretable model.
When an external force is applied, this separate component alters the system's dynamics, instead of altering the state directly.

The contribution of this paper is summarized as follows:
\begin{itemize}[leftmargin=5.5mm]
\item We propose a new framework \model for modeling interventions using Neural ODEs, where one component is dedicated to learning autonomous dynamics, while a separate module tracks the effect of external interventions. We provide specific examples of systems with different intervention types (\textit{e.g.}, permanent, decaying) and describe how \model can be implemented per different systems.
\item Using synthetic datasets and real-world medical records, we not only show \model consistently outperforms previous approaches for intervention modeling, but also analyze \model's behavior to show it is separately learning the autonomous dynamics and the intervention effects as intended.
\end{itemize}

\section{Background and Motivations}
\label{sec:background}
This section  briefly reviews Neural ODEs along with the definition of necessary notations.
We then motivate our work by discussing the limitation of existing approaches in intervention modeling. 

\vspace{1mm}
\noindent
\textbf{Neural ODEs}
The Neural ODEs provide a general framework for modeling the continuous transformation of the state (\textit{i.e.} latent representation $\hb$) by assuming the state dynamics can be modeled by an ordinary differential equation.
Neural ODEs parameterize the derivative of the state (\textit{i.e.} system dynamics ${d\hb}/{dt}$) with a neural network $f_{\theta}$ as 
\begin{equation}
    \dot\hb=f_\theta(t, \hb(t)),\quad \hb(t_{N})=\hb(t_{0})+\int_{t_{0}}^{t_{N}} f_\theta(\tau, \hb(\tau)) d\tau \label{eq:ode}
\end{equation}
Given an initial state vector $\hb(t_0)\in\R^{n_h}$, generally corresponding to the input vector $\xb\in\R^{n_x}$ or its embedding, the system state $\hb(t_k)$ is obtained by integrating forward the vector field $f_{\theta}:\R\times\R^{n_h}\times\R^{n_\theta}\rightarrow\R^{n_h}$, parameterized by $\theta\in\R^{n_\theta}$.
While we consider $\theta$ to be constant over time, the discussion below is directly compatible with time--varying parameters \cite{massaroli2020dissecting}.

\vspace{1mm}
\noindent
\textbf{Impulsive Systems}
Motivated by Neural ODEs' flexibility to model time-series data, several apporaches extended the Neural ODE framework to encode a sequence of observations $\{\xb_k\}$, whether being regular or irregular. 
ODE-RNN \cite{rubanova2019latent} uses the RNN to encode a sequence of observations, where the latent state $\hb$ flows according to an ODE between observations,
and a new observation $\xb_k$ directly modifies $\hb$ according to the RNN update equation.
GRU-ODE-Bayes \cite{de2019gru} specifically uses the GRU \cite{cho2014learning} to encode observations,
where a new observation directly modifies $\hb$ according to the GRU update equation combined with some masking operations.
Neural Jump Stochastic Differential Equations (NJSDE) \cite{jia2019neural} combines ODEs with point processes to model stochastic events.
In NJSDE, given a new observation, an event history representation is directly modified via an MLP, combining some internal state with the observation. 

Let $(\cT,\geq)$ be a finite linearly ordered set, called the \textit{data time} set (typically $\cT=\{t_0,t_1,\dots,t_N\}$).
We assume an \emph{input-output data stream} is given as a sequence $\{x_{t_k}\}_{t_k\in\cT}$.
For a compact notation, we denote $\xb_{t_k}$ with $\xb_k$.
The underlying core approach in all three approaches is to update the system state $\hb$ via a particular non-linear operation when a new observation is given, which can be generalized as an \textit{impulsive differential equation} \cite{lakshmikantham1989theory,kulev1988strong} of the type, 
\begin{equation}\label{eq:IDE}
    \left\{
    \begin{aligned}
        \dot \hb &= f_\theta(t, \hb(t)) &~~t\neq t_k\\
        \hb^+ &=  g_{\phi}(\hb(t), \xb_{k}) &~~t=t_k
    \end{aligned}
    \right.~~~~\text{for  } t_k\in\cT, 
\end{equation}
where $\hb^+$ indicates the value of $\hb$ after the discrete jump at $t_k$.
Between observations, the system state $\hb$ evolves according to continuous dynamics.
New observations, on the other hand, trigger a jump of $\hb$ to $\hb^+$ as determined by $g_{\phi}$.
For the rest of the paper, we refer to this broad family of models as \textit{Neural Jump Differential Equations} (NJDE).

Although it is possible to use NJDEs to naively model systems with interventions, by, say, simply concatenating them with observations, this approach is limited as it does not explicitly separate autonomous dynamics of the system from these effects. This, in turn, makes it challenging to properly reconstruct the evolution of a system subject to external forces.

It should be noted that according to Eq.~\ref{eq:IDE}, new inputs to NJSDEs induce state jumps. In general, states in dynamical systems are not guaranteed to jump in their entirety. For example, in mechanical systems with impacts, only higher--order states jump (\textit{e.g,} velocities).
Moreover, when the underlying dynamics is only partially observable, which is the case in numerous real-world problems (\textit{e.g.}, only 
a patient's temperature is measured), interventions cause jumps in a latent representation of the state, ultimately leading to an abrupt change in the dynamics driving the \textit{observable}. As a result, the state does not jump, but rather continuously changes according to the new, modified underlying dynamics.

\section{Proposed Approach}
\label{sec:method}
\begin{figure*}[t]
    \center
    \includegraphics[width=11cm]{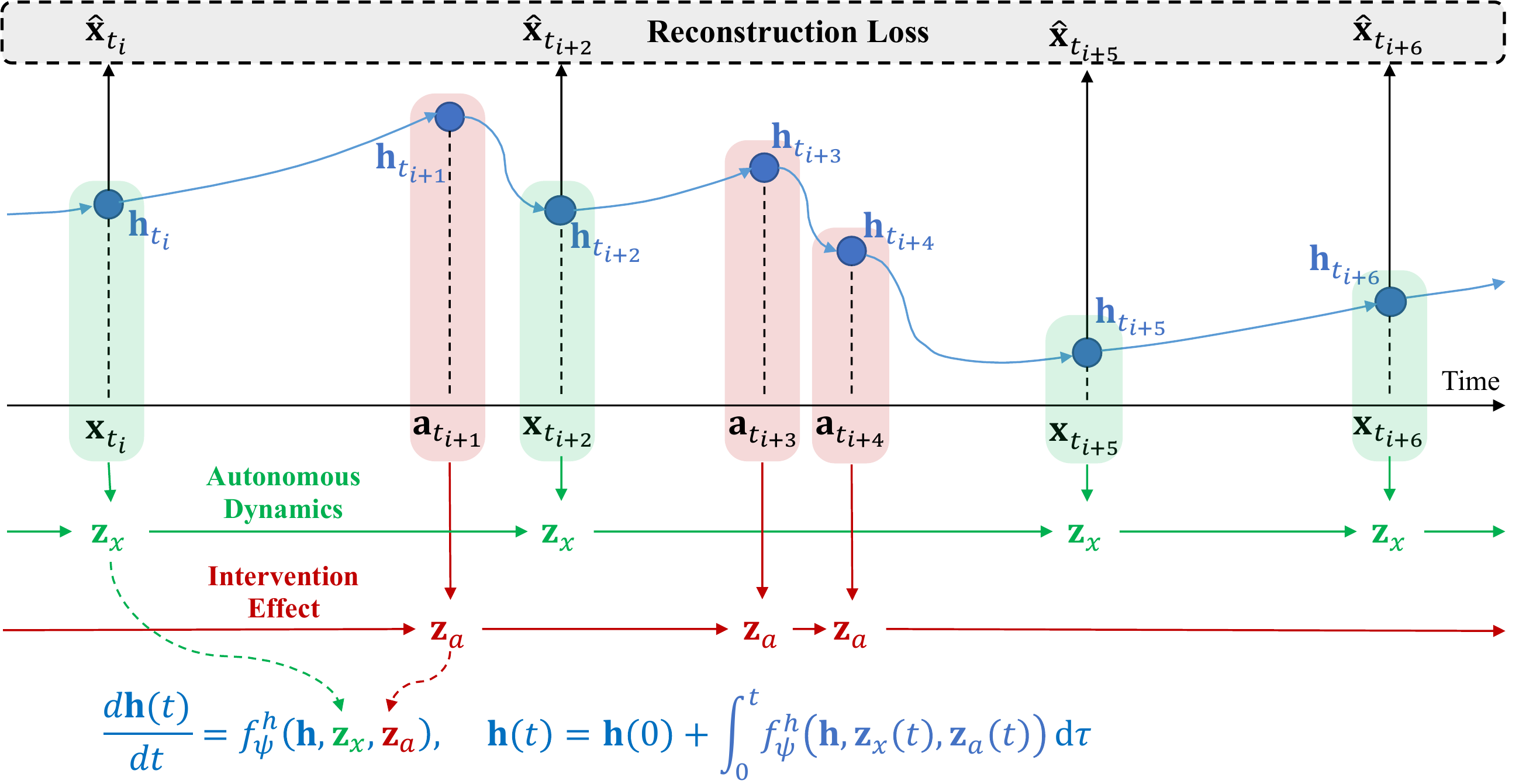}
    \caption{Overview of \model. $\zb_x$ represents the latent state of the autonomous dynamics based on observations $\xb$'s and states $\hb$'s, while $\zb_a$ represents the latent state of the intervention effect based on states $\hb$'s and actions $\ab$'s. $f^{h}_{\psi}$ combines $\zb_x$, $\zb_a$, and $\hb$ to obtain the final system dynamics $d\hb/dt$. We train our model via the reconstruction loss of the observations.}
    \label{fig:architecture}
\end{figure*}
With the objective of alleviating the underlying limitations in NJDE-based approaches, we propose \model, a novel framework designed to natively accommodate various types of intervention effects, common across application areas. \model can properly model both the system's autonomous dynamics as well as the effect of interventions.
We first describe the mathematical framework of \model, followed by the discussion on how it can be applied to different types of systems.

\subsection{Model Architecture of \model}
\label{ssec:architecture}
\textbf{Notation:}
We use $\xb_{t_k}$ to denote the observation at timestep $t_k$, and $\ab_{t_k}$ to denote an intervention at timestep $t_k$, where $t_k \in [0, T]$, $T$ being the end of the timeline.
Therefore an observation and an intervention can occur either at the same time (\textit{i.e.} $\xb_{t_k}, \ab_{t_k}$) or at different times.
We use $\Xb_{0:t_k}$ to denote all observations between time zero and $t_k$ (inclusive), and $\Ab_{0:t_k}$ to denote all interventions between time zero and $t_k$ (inclusive).
A comprehensive notation table for this section is provided in Appendix~\ref{appendix:notation}.

\vspace{1mm}
\noindent
\textbf{\model Framework:}
Figure \ref{fig:architecture} illustrates our model architecture at a high level. 
Observations $\Xb$ and interventions $\Ab$ occur in a particular order, and the system state $\hb$ evolves over time.
\model performs intervention modeling as 
\begin{equation}\label{eq:IMODE}
    \begin{matrix*}[l]
        &\dot \hb &= f^h_\psi(\hb, \zb_x, \zb_a)\\
        \text{continuous dynamics } & \dot \zb_x &= f^x_\theta(\zb_x) & t\neq t_k\\
        &\dot \zb_a &= f^a_\phi(\zb_a)\\[3pt]\hline\rule{0pt}{1.1\normalbaselineskip}
        &\hb^+ &= \hb\\
        \text{discrete dynamics }&\zb_x^+ &=  g^x_\theta(\hb, \zb_x, \xb_{t_k}) & t = t_k\\
        &\zb_a^+ &= g^a_\phi(\hb, \zb_a, \ab_{t_k})\\[3pt]\hline\rule{0pt}{1.1\normalbaselineskip}
        \text{predictor (decoder) }&\hat{\xb}(t) &= \ell_{\omega}(\hb(t)) 
    \end{matrix*}
\end{equation}
where $f^x_{\theta}, f^a_{\phi}$ model the independent vector fields of autonomous latent state $\zb_x$ and intervention effect $\zb_a$, respectively. Without external observations or interventions, the latent state $\hb$ evolves according to $f^h_\psi$, tasked with mixing instantaneous observation and intervention effects appropriately. This is opposed to standard Neural ODEs, where the dynamics is typically determined only by the latent system state $\hb$. 
\model is further equipped with specific components to incorporate sporadic observations and interventions, namely $g^x_{\theta}$ and $g^a_{\phi}$, which induce jumps on $\zb_x$ and $\zb_a$ while preserving continuity of $\hb$. It is often the case that the current and the past interventions have a combined effect on the system (\textit{i.e.} medications given to patients over time); to address such scenarios, we allow $g^x_{\theta}, g^a_{\phi}$ to leverage information contained in $\hb$, in addition to $\zb_x$ and $\zb_a$.

We train the various components of \eqref{eq:IMODE} with a reconstruction loss of the type:
\begin{equation}
\small
\cL := \frac{1}{K}\sum_{k = i}^K \|\xb_{t_k} - \hat\xb(t_k)\|_2^2 = \frac{1}{K}\sum_{k = i}^K \|\xb_{t_k} - \ell_\omega(\hb(t_k)\|_2^2 
\label{eq:recon_loss}
\end{equation}
where $\ell_\omega$ is a trainable decoding function that maps $\hb$ back to observation space $\xb$.
\model is therefore trained by solving the following nonlinear program
\begin{equation}\label{eq:3}
    \begin{aligned}
        \min_{\psi, \theta, \phi, \omega}  &\frac{1}{K}\sum_{k = i}^K \|\xb_{t_k} - \ell_\omega(\hb(t_k)\|_2^2\\
        \text{subject to}~~
        &\hb(t) =\hb(0) + \int_0^t f^h_\psi(\hb, \zb_x(\tau), \zb_a(\tau))\dd \tau\\
        & \begin{bmatrix}
        \dot \zb_x\\
        \dot \zb_a
        \end{bmatrix}
        =\begin{bmatrix}
        f^x_\theta(\zb_x)\\
        f^a_\phi(\zb_a)
        \end{bmatrix} \qquad~~~~~~~t\neq t_k\\
        & \begin{bmatrix}
        \zb_x^+\\
        \zb_a^+
        \end{bmatrix}
        =\begin{bmatrix}
        g^x_\theta(\hb, \zb_x, \xb_{t_k})\\
        g^a_\phi(\hb, \zb_a, \ab_{t_k})
        \end{bmatrix} ~~~t=t_k\\
        &~~ t \in [0, t_K]
    \end{aligned}
\end{equation}
\subsection{System--Specific Variants of \model}
\label{ssec:implementation}
The framework of our model allows a flexible implementation depending on the property of the target system.
In the following, we give a few concrete examples. 

\vspace{1mm}
\noindent
\textbf{Switching intervention effect:}
For example, a ball in uniform motion can be seen as having constant autonomous dynamics: given no external force, it will continue to move along the same course. However, if it comes in contact with another moving ball, its direction will permanently change. In these scenarios, appropriate architectural choices for model \eqref{eq:IMODE} would be, as an example, those provided in Table \ref{eq:ver1}.
\begin{table}[h]
    \centering
    \begin{tabular}{r|l||r|l}\toprule
        continuous dyn. & model& discrete dyn. & model \\\midrule
        $f^h_\psi$ & $\zb_x + \zb_a$ & $g^x_\theta$ & MLP\\
        $f^x_\theta$ & 0 & $g^a_\phi$ & MLP\\
        $f^a_\phi$ & 0 & $\ell_\omega$ & Id\\\bottomrule
    \end{tabular}
    \caption{\model variant for switching intervention effects.}
    \vspace{1mm}
    \label{eq:ver1}
\end{table}
With no intervention given, the autonomous vector field $f^x_\theta$ is solely determined by the current state $\hb_{t_i}$. If a collision occurs, the intervention latent state $\zb_a$ abruptly changes based on the colliding ball state, thus indirectly affecting $\hb$ through $f^h_\psi$. The intervention effect from $\zb_a$ is constant until a following collision happens; this, in turn, leads to a switching behavior of $f^h_\psi$ aligned with intervention events.

\vspace{1mm}
\noindent
\textbf{Decaying intervention effect:}
For example, we can imagine a patient in an Intensive Care Unit (ICU) whose cardiovascular function is slowly deteriorating.
Administering medications to this patient will have an effect that decays over time, as the ingredient is consumed by the system (\textit{i.e.} the patient).
Such a system can be modeled through the following component choices:
\begin{table}[h]
    \centering
    \begin{tabular}{r|l||r|l}\toprule
        continuous dyn. & model& discrete dyn. & model \\\midrule
        $f^h_\psi$ & MLP & $g^x_\theta$ & MLP\\
        $f^x_\theta$ & MLP & $g^a_\phi$ & MLP\\
        $f^a_\phi$ & $-\alpha\zb_a$ & $\ell_\omega$ & Id\\\bottomrule
    \end{tabular}
    \caption{\model variant for decaying intervention effects.}
    \vspace{3mm}
    \label{eq:ver2}
\end{table}

The patient's autonomous dynamics $f^x_{\theta}$ are kept general, and incorporate new observations through $g^x_{\theta}$. This, in turns, yields a combined effect on $\zb_x$ mimicking that of an ODE--RNN \cite{rubanova2019latent}. On the other hand, the latent intervention state $\zb_a$ is assumed to be decaying in time. We encode this prior information in the functional form of the flow $f^a_{\phi}$. 

\vspace{1mm}
\noindent
\textbf{Generalized Implementation:}
Based on two concrete examples above, we propose the most general form of \model that can ideally model any systems with irregular observations and interventions, without assuming prior knowledge of the autonomous dynamics and the intervention effect. Model \eqref{eq:ver3} does not assume any particular functional form of intervention effects and system dynamics, making it a particularly appropriate general purpose choice.

\begin{table}[h]
    \centering
    \begin{tabular}{r|l||r|l}\toprule
        continuous dyn. & model& discrete dyn. & model \\\midrule
        $f^h_\psi$ & MLP & $g^x_\theta$ & MLP\\
        $f^x_\theta$ & MLP & $g^a_\phi$ & MLP\\
        $f^a_\phi$ & MLP & $\ell_\omega$ & MLP\\\bottomrule
    \end{tabular}
    \caption{\model variant for general purpose settings.}
    \vspace{4mm}
    \label{eq:ver3}
\end{table}

\section{Experiments}
\label{sec:exp}
We evaluate \model with two simulated datasets and one real-world medical records both quantitatively and qualitatively.
The two simulated datasets (Moving Ball and Exponential Decay) represent systems with permanent-effect interventions and decaying-effect interventions, respectively.
We also use eICU~\cite{pollard2018eicu}, a publicly available electronic health records, to evaluate \model's performance on real-world data.
Detailed experimental settings including the hyperparameters
are described in Appendix~\ref{appendix:hyperparams}.
All datasets and \model source code are available at GitHub\footnote{\url{https://github.com/eogns282/IMODE}}.

\subsection{Moving Ball \& Exponential Decay}
\label{ssec:ball_decay}
\begin{figure*}[t]
    \center
    \includegraphics[width=13.9cm]{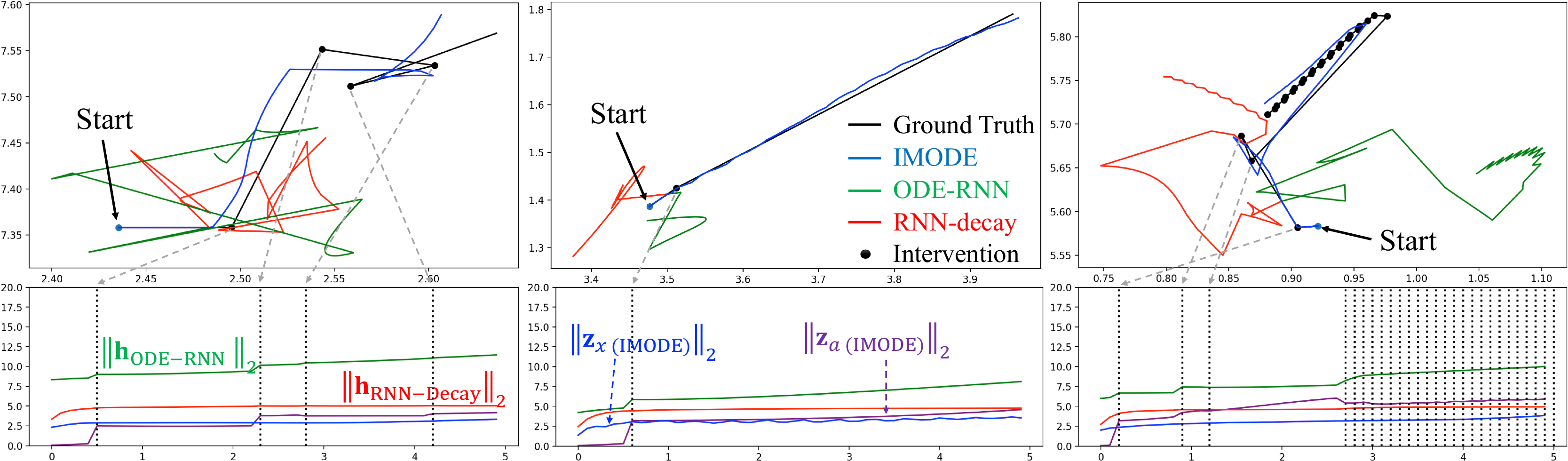}
    \caption{Simulated trajectories and $L_2$ norms of latent states of \model and baselines for the Moving Ball dataset. The gray dotted lines connects collision points to the corresponding timesteps. The three samples represent medium, light, and heavy interventions, respectively. }
    \label{fig:collision_viz}
\end{figure*}

We use two simulated datasets, \textit{Moving Ball} and \textit{Exponential Decay} to demonstrate \model's capability to model the first two cases described in Section \ref{ssec:implementation}: permanent-effect interventions, and changing autonomous dynamics with time-decaying effects of intervention.
Observations $\xb$ and interventions $\ab$ in Moving Ball consist of 2D positions of the target ball, and 2D positions and 2D velocities of intervening balls, sampled from a contact simulator\footnote{\url{https://scipython.com/blog/two-dimensional-collisions}}, where we set all balls to have the same size and mass.
Example trajectories can be seen in Figure~\ref{fig:collision_viz}.
Note that the velocity of the target ball is unobserved, forcing the models to infer the velocity based on positions.

\begin{figure*}[t]
    \center
    \includegraphics[width=13.9cm]{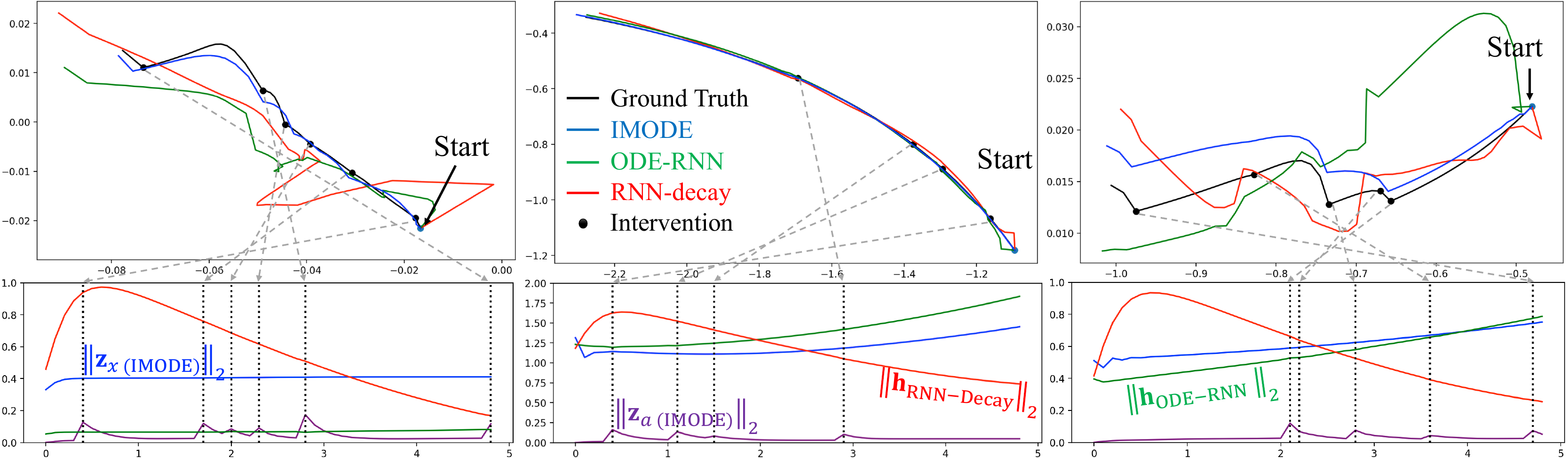}
    \caption{Simulated trajectories and $L_2$ norms of latent states of \model and baselines for the Exponential Decay dataset. The gray dotted lines connects intervention points to the corresponding timesteps. The three samples represent medium, light, and heavy interventions, respectively. }
    \vspace{-2mm}
    \label{fig:synthetic_viz}
\end{figure*}

In Exponential Decay, observations $\xb$ consist of 2D positions that follow a deterministic dynamics based on $\xb$.
Interventions $\ab$ consist of 2D values sampled from $\mathcal{N}(\mathbf{0, I})$ with $10\%$ chance at every time unit.
The effect of interventions is determined by a non-linear function (\textit{i.e.} a randomly initialized MLP with ReLU activation) using $\xb, d\xb/dt$ and $\ab$ as input.
The effect is added to $\xb$, and the value of an effect is halved at every time unit.
Example trajectories can be seen in Figure~\ref{fig:synthetic_viz}.
Note that the intervention effect is hidden from the model, as only the randomly sampled $\ab$ is given.
We describe the simulation algorithm of Exponential Decay in Appendix~\ref{appendix:exp_dec_algorithm}.
In both Moving Ball and Exponential Decay, the initial position $\xb_0$ and the initial velocity $d\xb_0/dt$ are randomly chosen.
For both Moving Ball and Exponential Decay, we generated $1,000$ samples for training, $100$ for validation, and $100$ for testing, where all samples have the sequence length of $50$.

\begin{table*}[t]
\begin{center} \small 
\begin{tabular}{lcccc}
\toprule
\multicolumn{1}{c}{\multirow{2}{*}{\textbf{Methods}}}
& \multicolumn{2}{c}{\textbf{Moving Ball ($ \times 10^{-2}$)}} & 
\multicolumn{2}{c}{\textbf{Exponential Decay ($ \times 10^{-4}$)}}\\
\cmidrule(lr){2-3} \cmidrule(lr){4-5}
 & Validation MSE & Test MSE & Validation MSE & Test MSE\\
\midrule
GRU-$\Delta_t$ & 6.033 ($\pm$ 0.272) & 5.994 ($\pm$ 0.102) & 4.381 ($\pm$ 2.004) & 5.686 ($\pm$ 3.258)\\
GRU-Decay & 6.384 ($\pm$ 0.059) & 5.994 ($\pm$ 0.102) & 5.135 ($\pm$ 1.660) & 6.589 ($\pm$ 5.109)\\
ODE-RNN & 2.478 ($\pm$ 0.142) & 2.502 ($\pm$ 0.328) & 3.342 ($\pm$ 1.161) & 2.778 ($\pm$ 1.173)\\
GRU-ODE-Bayes & 2.506 ($\pm$ 0.436) & 2.597 ($\pm$ 0.520) & 3.852 ($\pm$ 2.903) & 3.643 ($\pm$ 4.996)\\
CRN & 2.209 ($\pm$ 0.065) & 2.220 ($\pm$ 0.299) & 3.716 ($\pm$ 2.542) & 4.881 ($\pm$ 3.846)\\
\midrule
\model$_{switch}$ & 1.914 ($\pm$ 0.173) & \textbf{1.794} ($\pm$ 0.203) & \textbf{0.019} ($\pm$ 0.001) & \textbf{0.027} ($\pm$ 0.005)\\
\model$_{decay}$ & \textbf{1.794} ($\pm$ 0.203) & 1.816 ($\pm$ 0.203) & 0.142($\pm$ 0.084) & 0.131 ($\pm$ 0.490)\\
\model$_{general}$ & 1.798 ($\pm$ 0.215) & 1.824 ($\pm$ 0.230) & 0.039 ($\pm$ 0.009) & 0.041 ($\pm$ 0.008)\\
\bottomrule
\end{tabular}
\caption{Validation and test MSE of all models on Moving Ball and Exponential Decay.}
\label{tab:bb_ed_result}
\end{center}
\end{table*}
\vspace{1mm}
\noindent
\textbf{Baseline Methods}
We compare \model against both RNN-based and ODE-based methods:
GRU with time-gap information (GRU-$\Delta_t$),
GRU with exponentially decaying hidden states (GRU-Decay),
ODE-RNN \cite{rubanova2019latent},
and GRU-ODE-Bayes \cite{de2019gru}.
Specifically, GRU-decay can be seen as using prior knowledge of intervention effects, similar to classical intervention models discussed in Section~\ref{sec:related}.
Note that observations $\xb$ occur regularly (\textit{i.e.} no missing value) but the interventions occur irregularly.
To feed observations $\Xb$ and interventions $\Ab$ to the baseline models, we align both time-series data by the time and concatenate them.
If $\ab$ does not exist at some timestep $t_k$, we set it to a zero vector.
Given input $[\Xb_{0:t_k};\Ab_{0:t_k}]$, RNN-based models are trained to predict $\xb_{t_{k+1}}$.

ODE-based models are trained in the same fashion as \model (Eq.~\ref{eq:recon_loss}).
We also test three variants of \model differing in terms of expressiveness: 1) \model$_{switch}$ from Table~\ref{eq:ver1}, 2) \model$_{decay}$ from Table~\ref{eq:ver2}, 3) \model$_{general}$ from Table~\ref{eq:ver3}.
We also use Counterfactual Recurrent Network (CRN) \cite{bica2020estimating} as a baseline, the state-of-the-art model that takes interventions into account when modeling observations. CRN, however, models only discrete interventions whereas interventions in Moving Ball and Exponential Decay are continuous. We therefore use a modified CRN (modification details are provided in Appendix~\ref{appendix:crn}).

\vspace{1mm}
\noindent
\textbf{Quantitative Evaluation}
During training, we fed the first $10$ true observations (\textit{i.e.} the 2D positions) to each model, and then made it evolve for the remaining $40$ steps, while always using true intervention values $\Ab$ from time $0$.
Model parameters were updated via the MSE loss between the predicted observations $\hat{\Xb}$ and the true observations $\Xb$. 
The test MSEs were measured in the same fashion; given the first $10$ true positions, and the true intervention information, simulate the remaining $40$ steps.
We conduct 5-fold cross-validation for all experiments.

As seen in Table~\ref{tab:bb_ed_result}, all \model variants consistently outperform the baseline models for both datasets.
Moreover, \model$_{general}$ shows robust performance in both Moving Ball and Exponential Decay, demonstrating its capability to learn two significantly different dynamical systems.
The performance gap between baselines and \model is much larger for Exponential Decay, indicating that \model is a suitable framework especially in modeling the system with a global pattern (\textit{i.e.} autonomous dynamics) and local perturbations (\textit{i.e.} interventions).
It is also noteworthy that the MSEs of all models are shown to be significantly higher for Moving Ball compared to Exponential Decay, probably due to the difficulty of modeling acute changes in the ball dynamics, as seen in Figure~\ref{fig:collision_viz}.

\begin{figure*}[t]
    \center
    \includegraphics[width=13.5cm]{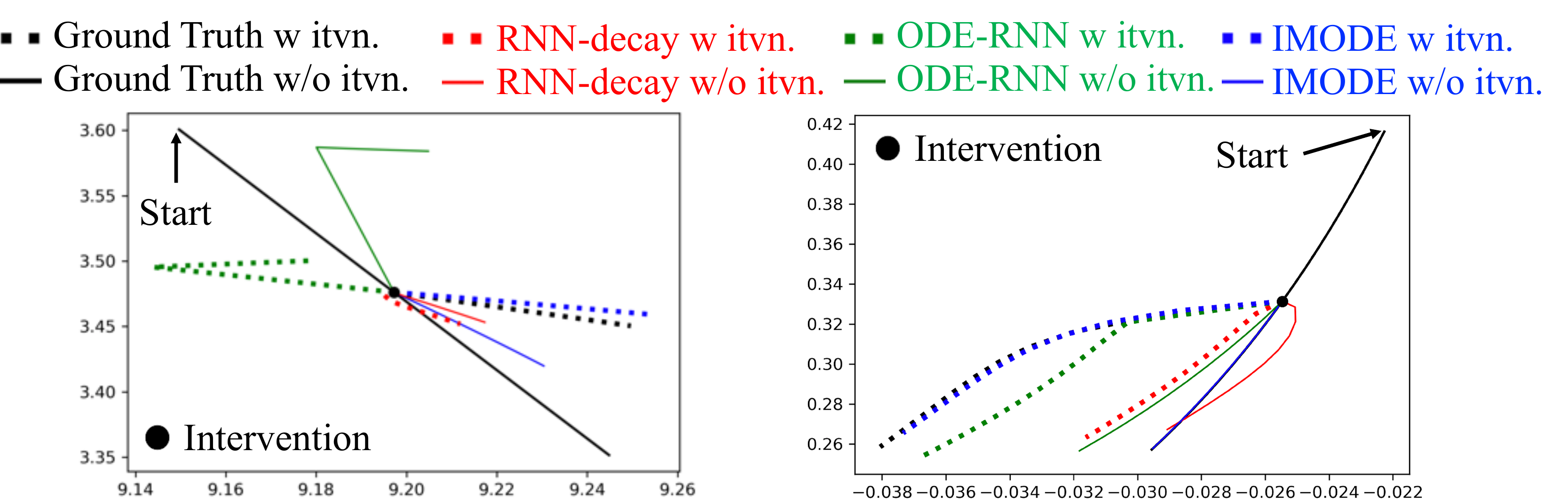}
    \caption{
    (\textit{Left}) Simulated trajectories of \model and baselines for Moving Ball. 10 steps are simulated before trajectory divides into two alternative futures.
    (\textit{Right}) Simulated trajectories of \model and baselines for Exponential Decay. In the same manner, 10 steps are simulated before trajectory divides into two alternative futures.
    }
    \label{fig:cf_viz}
\end{figure*}

\begin{table}[t]
\begin{center}
\resizebox{\columnwidth}{!}{%
\begin{tabular}{lcc}
\toprule
 Methods & Moving Ball ($\times 10^{-2}$) & Exponential Decay ($\times 10^{-4}$)\\
\midrule
GRU-$\Delta_t$ & 1.761($\pm$ 0.204) & 2.014 ($\pm$ 0.650)\\
GRU-Decay & 1.802 ($\pm$ 0.184) & 2.289 ($\pm$ 0.375)\\
ODE-RNN & 1.280 ($\pm$ 0.179) & 1.771 ($\pm$ 0.654)\\
GRU-ODE-Bayes & 1.318 ($\pm$ 0.064) & 1.246 ($\pm$ 0.451)\\
CRN & 0.682 ($\pm$ 0.205) & 1.446 ($\pm$ 0.498)\\
\midrule
\model$_{switch}$ & 0.247 ($\pm$ 0.017) & 0.092 ($\pm$ 0.002) \\
\model$_{decay}$ & 0.222 ($\pm$ 0.035) & \textbf{0.082} ($\pm$ 0.003) \\
\model$_{general}$ & \textbf{0.221} ($\pm$ 0.035) & 0.094 ($\pm$ 0.003)\\
\bottomrule
\end{tabular}
}
\caption{Test MSEs of all models in counterfactual scenarios (alternative futures) using two datesets.}
\label{tab:cf_result}
\end{center}
\end{table}
We further tested all models in counterfactual scenarios using both datasets where a single trajectory, after 10 initial steps, divides into two alternative futures (with and without an intervention) and continues for another 10 steps.
Example trajectories can be seen in Figure~\ref{fig:cf_viz}.
We feed the first 10 steps to the already trained models from Table~\ref{tab:bb_ed_result} and then let them simulate the next 10 steps for two alternative futures.
As seen from Table~\ref{tab:cf_result}, \model variants again outperform all baselines in these counterfactual scenarios.
The fact that \model is able to separately learn autonomous dynamics and intervention effects clearly indicates its capability to generalize to alternative cases.

\vspace{1mm}
\noindent
\textbf{Model Behavior Analysis}
To confirm that \model properly learns the autonomous dynamics and the intervention effect, we visualize test trajectories from Moving Ball (Figure~\ref{fig:collision_viz}) and Exponential Decay (Figure~\ref{fig:synthetic_viz}), where the first $10$ true timesteps are given to the model, and the model simulates the remaining $40$ steps while using true interventions. 
We show the results of \model$_{general}$, ODE-RNN, and RNN-Decay for comparison.
In both figures, \model clearly outperforms baseline models as it closely follows the true trajectories, while the baselines often diverge.
The comparison between Figures~\ref{fig:collision_viz} and \ref{fig:synthetic_viz} demonstrates the challenging nature of Moving Ball, thus resulting in the higher MSE in Table~\ref{tab:bb_ed_result}.
Whereas Exponential Decay shows smooth and moderate change of dynamics, the changes in Moving Ball are not only discrete but also significant (\textit{e.g.}, ball changing direction in almost 180 degrees).

Figure~\ref{fig:collision_viz} shows that the autonomous latent state $\zb_x$ stays rather static, while the intervention latent state $\zb_a$ \textit{jumps} when a collision occurs.
One can also see from this figure that after the jump, $\zb_a$ does not decay over time, indicating that \model is successfully recording the permanent effects of all the previous interventions.
Ideally, $\zb_x$ should remain constant over time, but minor changes and fluctuations are found, especially in the second trajectory of Figure~\ref{fig:collision_viz}, leaving room for further improvement.
Based on the trajectories of $||\hb_{\text{ODE-RNN}}||_2$, ODE-RNN also recognizes the occurrence of collisions, but it fails to treat observations and interventions separately, leading to incorrect simulation. \\
Compared to Moving Ball, Exponential Decay is a completely different system where the state follows its own dynamics while being occasionally perturbed with an exponentially decaying effect.
Thanks to its relatively smooth trajectory, even the baseline models tend to stay close to the true trajectory.
However, the $L_2$ norm plots of Figure~\ref{fig:synthetic_viz} clearly demonstrates the benefit of separately modeling observations and interventions.
While $||\zb_x||_2$ demonstrates a smoothly changing trajectory potentially corresponding to the autonomous system dynamics, $||\zb_a||_2$ jumps when an intervention occurs and decays over time, indicating that \model has properly learned the true intervention effect on the system.

We also provide visual examples for the counterfactual scenarios.
The two figures in Figure~\ref{fig:cf_viz} describe two counterfactual cases in Moving Ball and Exponential Decay respectively.
As indicated by the quantitative results in Table~\ref{tab:cf_result}, \model outperforms baseline models in both datasets, as it closely follows the two alternative futures while the baselines diverge from the true trajectories in both alternative cases.
\vspace{-2mm}
\subsection{eICU Dataset}
\vspace{-0.5mm}
The eICU Collaborative Research Database (eICU)~\cite{pollard2018eicu} contains publicly available electronic health records (EHR) collected from multiple intensive care units (ICU).
In order to correctly evaluate \model's ability to learn the patient's autonomous dynamics and the effect of interventions (\textit{i.e.} drugs), we choose a patient with the longest ICU stay whose drugs were given only via IV infusion to remove any confounding factors (\textit{e.g.}, drugs taken orally).
We focus on a single patient since every patient has a unique autonomous dynamics and response to drugs determined by hidden factors (\textit{e.g.}, DNA and diet).
We leave handling multiple heterogeneous dynamics with a single model as future work.
We extract from the EHR three blood pressure features (systolic, diastolic, and mean) measured every 5 minutes as the observation $\xb$, and the hourly interventions $\ab$ consist of five drug types (norepinephrine, vasopressin, propofol, amiodarone, and  phenylephrine) and their dosage.
We binned the entire observations into 2.5-hour buckets (30 timesteps containing two interventions) and used 150 buckets for training, 50 for validation and 50 for testing.

\vspace{-2mm}
\noindent
\textbf{Quantitative Evaluation}
\begin{table}[t]
\begin{center}
\resizebox{\columnwidth}{!}{%
\begin{tabular}{lcc}
\toprule
 Methods & Validation MSE ($\times 10^{-3}$) & Test MSE ($\times 10^{-3}$)\\
\midrule
GRU-$\Delta_t$ & 6.010 ($\pm$ 0.337) & 6.061($\pm$ 1.099)\\
GRU-Decay & 6.789 ($\pm$ 0.374) & 6.899 ($\pm$ 0.445)\\
ODE-RNN & 4.234 ($\pm$ 0.337) & 4.414 ($\pm$ 0.412)\\
GRU-ODE-Bayes & 5.588 ($\pm$ 1.258) & 5.988 ($\pm$ 1.350)\\
CRN & 5.551 ($\pm$ 1.193) & 5.771 ($\pm$ 1.125)\\
\midrule
\model$_{switch}$ & 6.437 ($\pm$ 3.332) & 6.410 ($\pm$ 3.056) \\
\model$_{decay}$ & 4.262 ($\pm$ 0.355) & 4.245 ($\pm$ 0.250) \\
\model$_{general}$ & \textbf{4.047} ($\pm$ 0.255) & \textbf{4.209} ($\pm$ 0.308)\\
\bottomrule
\end{tabular}
}
\caption{Validation and test MSE of all models on eICU.}
\label{tab:eicu_result}
\end{center}
\end{table}
Using the same set of baselines as in the above experiments, we train all models with the reconstruction loss (Eq.~\ref{eq:recon_loss}) in a similar fashion as before; $6$ true timesteps are given to the models, and the remaining $24$ steps are simulated using true intervention information.
We conduct 5-fold cross validation for all models.
As can be seen in Table~\ref{tab:eicu_result}, \model shows the best test performance, demonstrating its potential applicability to real-world data such as patient vital signs.
We also present further analysis of the model behavior in Appendix~\ref{appendix:eicu_viz}.

\section{Related Work}
\label{sec:related}
\textbf{Continuous--Depth Learning}
Continuous--depth learning \cite{sonoda2017double,haber2017stable,hauser2017principles,lu2017beyond,che2018recurrent,massaroli2020dissecting} has recently emerged as a novel paradigm providing a dynamical system perspective on machine learning. This view has inspired design of novel architectures \cite{chang2017multi, zhu2018convolutional,demeester2019system,chang2019antisymmetricrnn,cranmer2020lagrangian,massaroli2020stable} as well as guiding the injection of physics--inspired inductive biases \cite{greydanus2019hamiltonian,kohler2019equivariant}. The framework has seen applications to various classes of differential equations \cite{tzen2019neural,li2020scalable} and graphs \cite{poli2019graph}, along with several analyses regarding computational speedups through regularization \cite{finlay2020train} or specific numerical methods \cite{poli2020hypersolvers}.

\cite{rubanova2019latent} demonstrated promising empirical performance across various forecasting datasets by combining RNNs with Neural ODEs. \cite{yildiz2019ode2vae} refined the architecture through higher--order dynamics and Bayesian networks.
\cite{de2019gru} alternates a filtering and predictions steps to improve performance in settings with highly sporadic observations.
\cite{jia2019neural} models a stochastic event by estimating the occurrence probability with Neural ODEs, where a new event observation updates the event intensity.
While the above approaches share similarities with the proposed approach such that a input sequence modifies the internal state, they fail to treat observations and interventions differently, leading to suboptimal performance when modeling external interventions in a given system.

\noindent
\textbf{Intervention Modeling}
Intervention modeling is typically discussed in the context of time-series analysis.
Combining the intervention analysis technique with the classical time-series models enables the user to handle time-series data with different types of interventions such as permanent, gradually increasing or decreasing, and complex effects~\cite{glass2008design}.
Intervention analysis has been used across diverse domains such as healthcare~\cite{evans2002effectiveness, wagner2002segmented},  economics~\cite{box1975intervention} and policies~\cite{enders1993effectiveness}.
More recent studies have been conducted in the context of patient modeling, where models based on a Gaussian Process (GP) and RNNs have been proposed \cite{schulam2017reliable, soleimani2017treatment, lim2018forecasting, bica2020estimating}.
Considering the restricted model structure that GP assumes, we used the state-of-the-art patient modeling algorithm from \citeauthor{bica2020estimating} as one of the baselines.
Causal analysis is also relevant to our work, but it aims to identify the causal relationship between input (usually a mixture of causal factors and confounders) and output~\cite{pearl2009causality}.
On the other hand, intervention modeling focuses on correctly predicting the effect of an external force. As intervention modeling is essentially a time-series problem, Neural ODEs are a natural framework for this task.

\section{Conclusions}
\label{sec:conclusion}
In this work, we proposed \model, a Neural ODE-based framework that can properly model dynamical systems with external interventions.
\model employs two components where one models the autonomous dynamics of the system, and the other models the intervention effect on the system.
Using both simulated and real-world datasets, we quantitatively demonstrated \model's superiority in intervention modeling, as well as in-depth analysis on its behavior. 
As future work, we plan to apply \model in large-scale real-world datasets while extending \model to further disentangle the autonomous dynamics and the intervention effects.

\section*{Ethical Impact}
\label{sec:impact}
Although we empirically demonstrated that the proposed framework \model is capable of learning separate latent states for both autonomous dynamics and intervention effects, it should be used with caution in real-world applications.
As described in Section~\ref{sec:related}, \model is not a causal analysis model, which means that the user must possess domain knowledge as to which variables are observations and which are interventions, and that there are no unobserved confounders that can affect the given dynamical system (as described in the eICU Dataset subsection).
For example, we believe \model can be used to model patient status in a well-controlled environment such as patients under anesthesia during operation.
We are certain more opportunities will follow as we address issues such as scalability and confounding factors in the future.

\small
\bibliography{reference}

\newpage
\clearpage
\appendix
\section{Notation Table}
\label{appendix:notation}
\small
\begin{table*}[t]
    \centering
    \begin{tabular}{r|l||l}\toprule
        Symbol & Description & Domain (and codomain) \\\midrule
        $\xb$ & input & $\R^\nX$\\
        $\ab$ & intervention & $\R^\nA$\\
        $\hb$ & continuous latent state & $\R^\nH$\\
        $\zb_x$ & autonomous latent state & $\R^\nZX$\\
        $\zb_a$ &  intervention latent state & $\R^\nZA$\\
        $\psi$ &  $\hb$'s dyn. parameters & $\R^\nPs$\\
        $\theta$ &  $\zb_x$'s dyn. parameters & $\R^\nTh$\\
        $\phi$ &  $\zb_a$'s dyn. parameters & $\R^\nPh$\\
        $\omega$ &  decoder's parameters & $\R^\nOm$\\
        $f^h_\psi$ &  $\hb$'s flow map & $\R^\nH\times\R^\nZX\times\R^\nZA\times\R^\nPs\rightarrow\R^\nH$\\
        $f^x_\theta$ &  $\zb_x$'s flow map & $\R^\nH\times\R^\nZX\times\R^\nTh\rightarrow\R^\nZX$\\
        $f^a_\theta$ & $\zb_a$'s flow map & $\R^\nH\times\R^\nZA\times\R^\nPh\rightarrow\R^\nZA$\\
        $g^x_\theta$ & $\zb_x$'s jump map &  $\R^\nH\times\R^\nZX\times\R^\nX\times\R^\nTh\rightarrow\R^\nZX$\\
        $g^a_\phi$ & $\zb_a$'s jump map & $\R^\nH\times\R^\nZA\times\R^\nA\times\R^\nPh\rightarrow\R^\nZA$\\
        $\ell_\omega$ & output decoder & $\R^\nH\times\R^\nOm\rightarrow\R^\nX$\\ 
        \bottomrule
    \end{tabular}
    \caption{Notations used throughout the paper.}
    \label{tab:my_label}
\end{table*}
Table~\ref{tab:my_label} describes the notations and their descriptions used throughout this paper.

\section{Hyperparameters}
\label{appendix:hyperparams}
In the experiments section, we evaluated five baseline methods: RNN-$\Delta_t$, RNN-Decay, ODE-RNN, GRU-ODE-Bayes and CRN.
We trained all baseline models for the same number of epochs in each experiment, and the final models were chosen by the validation loss in each epoch.
Additionally, we trained all ODE-based models using the Runge-Kutta fourth-order method and the same delta-times.

\vspace{1mm}
\noindent
\textbf{Hyperparameters for \model}\hspace{2mm}
In the Moving Ball task, we used the batch size of 32 for 1,000 epochs. 
When using RNNs or ODE-RNNs for both $f^x_\theta$ and $f^a_\phi$, the size of the hidden vector was 40.
Additionally, when using ODE-RNNs for $f^x_\theta$ and $f^a_\phi$, their derivative functions were a two-layer MLP with the hidden size of 40 and LeakyReLU as the activation function.
For $f^h_\psi$, we used the same 40-dimensional two-layer MLP with LeakyReLU activation.
We used the RMSprop optimizer with the learning rate of 0.001 and set the delta-time as $0.01$.
In Exponential Decay, we used the same setting as Moving Ball but used 1500 epochs.

In the eICU task, we trained for 1,500 epochs with the batch size of 32.
We used ODE-RNNs for the functions $f^x_\theta$ and $f^a_\phi$ with 20-dimensional and 10-dimensional hidden vectors respectively.
For $f^h_\psi$, we used the 20-dimensional two-layer MLP with LeakyReLU activation.
We used the delta-time of $1.0$ for the ODE solver.

\vspace{1mm}
\noindent
\textbf{Baselines}\hspace{2mm}
For the baselines in our experiments, 
we follow the general structure and hyperparameters of each model’s available implementation\footnote{\url{https://github.com/YuliaRubanova/latent_ode}}\textsuperscript{,}\footnote{\url{https://github.com/edebrouwer/gru_ode_bayes}}\textsuperscript{,}\footnote{\label{grud_github}\url{https://github.com/zhiyongc/GRU-D}}\textsuperscript{,}\footnote{\url{https://github.com/ioanabica/Counterfactual-Recurrent-Network}} other than small details.
Specifically, the models were tuned by the performance in the validation phase in order to obtain the proper batch size and learning rate.
We also adjusted the hidden vector dimension of the baseline models to match that of \model.

\section{Algorithm of Exponential Decay}
\label{appendix:exp_dec_algorithm}
\begin{algorithm}
   \caption{The simulation algorithm of Exponential Decay}
   \label{alg:Exponential Decay Gen}
   
\begin{algorithmic}[1]
   \STATE {\bfseries Input:} time unit $dt$, length of time series $K$
    
   \STATE Initialize observation $\xb_0$, $d\xb_0/dt$ are randomly chosen in [0, 1], intervention effect $e_{0}$ = 0, \newline update matrix of dynamics $(d\xb/dt)$ $M_{v}=\begin{bmatrix}
                1.5 & 0 \\
                0 & -2.5
                \end{bmatrix}$

   \FOR{$k=1$ {\bfseries to} $K $}
       \STATE $a_{k}=0$\;
       \STATE $\xb_{k} = \xb_{k-1} + dt*(d\xb_{k-1}/dt + e_{k-1})$
       \STATE $d\xb_{k}/dt = M_{v}(d\xb_{k-1}/dt)$  
       \STATE $e_{k} = e_{k-1}*0.5 $        
       
       \STATE $ intervention\_occurs \sim Bernoulli(0.1)$  
       \IF{$intervention\_occurs$}
           \STATE $a_{k} \sim \mathcal{N}(0, 1)$\;
            \STATE $e_{k} = e_{k} + MLP([\xb_{k}, d\xb_{k}/dt, a_{k}])$
       \ENDIF
   \ENDFOR
\RETURN{$\Xb_{1:K}, \Ab_{1:K}$}

\end{algorithmic}
\end{algorithm}
Algorithm~\ref{alg:Exponential Decay Gen} describes the pseudo-code for generating simulated samples used in the Exponential Decay task.

\section{Modification to Counterfactual Recurrent Network}
\label{appendix:crn}
Counterfactual Recurrent Network (CRN) (Bica et al. 2020) has Gradient Reversal Layer (GRL) that suppresses the correct prediction of the treatment type that occurs in the next timestep.
This technique cannot be used in Moving Ball and Exponential Decay, because the interventions in those datasets consist of continuous values (\textit{i.e.}, position and velocities of the incoming ball in Moving Ball, and the randomly generated intervention effect in Exponential Decay).
Although the GRL component is able to predict \textit{No Treatment} class as well, since the interventions occur randomly in Moving Ball and Exponential Decay, predicting the binary case of \textit{Treatment} and \textit{No Treatment} cannot be done either.
Therefore, in the two simulated datasets (Moving Ball and Exponential Decay) that have continuous and unpredictable interventions, we used CRN with $\lambda=0$ (\textit{i.e.,} not using a treatment classifier and GRL).

In the eICU experiment, although the value of intervention is continuous (\textit{i.e.}, the dosage of each treatment), their treatment type and its administration time would be predictable using observational trajectories of patient. Therefore, as the original setting in CRN, we used a treatment classifier to predict treatment types excluding their dosage. Additionally, since the patients can be given multiple treatments simultaneously in the eICU experimental settings, we used the sigmoid activation function in the last layer of the treatment classifier instead of softmax function to predict the multiple treatments at the same time.

\section{Model Behavior Analysis for eICU}
\label{appendix:eicu_viz}

\begin{figure*}[h]
    \center
    \includegraphics[width=13cm]{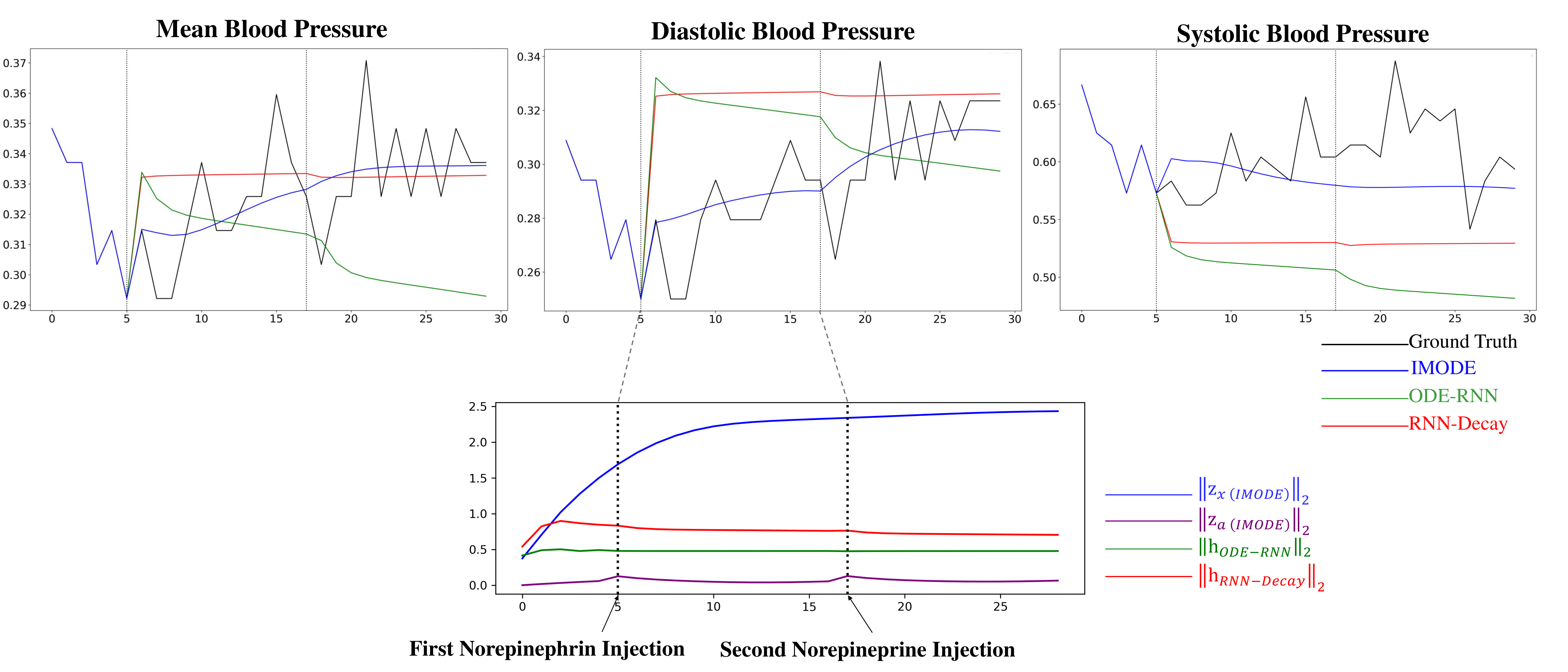}
    \caption{Three trajectories (mean blood pressure, diastolic blood pressure, and systolic blood pressure) and $L_2$ norms of hidden layers of \model and baselines for the eICU dataset. The gray-dotted lines connect intervention points (\textit{i.e.} when norepineprine was injected) to the corresponding timesteps.}
    \label{fig:eicu_02}
\end{figure*}

\begin{figure*}[h]
    \center
    \includegraphics[width=13cm]{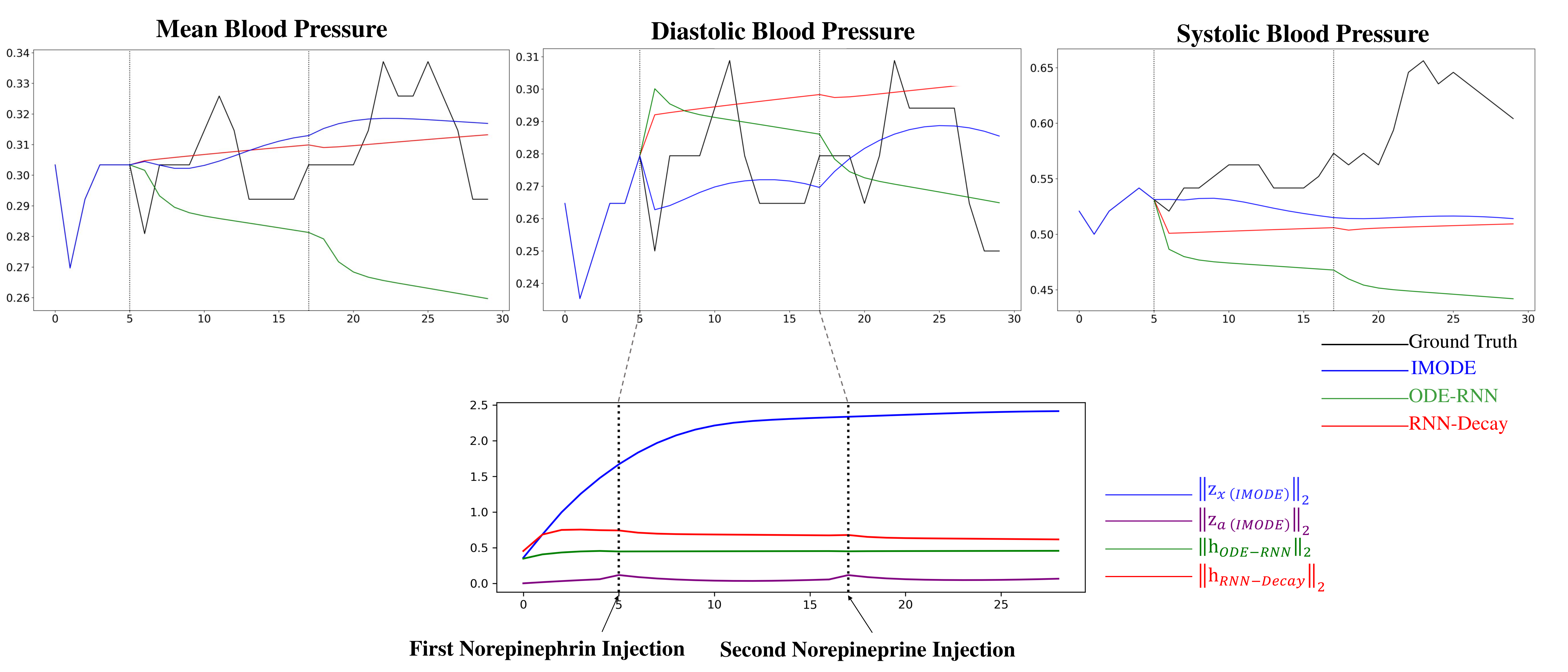}
    \caption{Another set of three trajectories (mean blood pressure, diastolic blood pressure, and systolic blood pressure) and $L_2$ norms of hidden layers of \model and baselines for the eICU dataset. The gray-dotted lines connect intervention points (\textit{i.e.} when norepineprine was injected) to the corresponding timesteps.}
    \label{fig:eicu_03}
\end{figure*}

In this section, we visualize the trajectories of eICU samples and their $L_2$ norms to study the model behaviors.
In Figure~\ref{fig:eicu_02}, the model was given true observations and interventions for the first six steps.
Then, each model autoregressively predicted the observations for the remaining 24 timesteps while using true interventions.

As depicted in both Figure~\ref{fig:eicu_02} and \ref{fig:eicu_03}, the blood pressure features seem to increase after the injection of norepinephrine which has a direct influence to rise of blood pressure, but the exact effect varies by the patient. \model most accurately predicts the blood pressure trajectories and effects of treatments.
We found systolic blood pressure to be more unpredictable than mean blood pressure and diastolic blood pressure (for all models), as they demonstrated seemingly random trajectories, suggesting that there exist other unknown factors affecting the patients besides the medications.

As can be seen from the $L_2$ norm activities in both figures, \model successfully disentangles the autonomous dynamics of the patient and the intervention effect, where $||\zb_x||_2$ demonstrates a rather stable trajectory while $||\zb_a||_2$ spikes when there are interventions followed by a gradual decay.

\section*{References}
Bica, I.; Alaa, A. M.; Jordon, J.; and van der Schaar, M. 2020. Estimating  counterfactual treatment outcomes over time through adversarially balanced representations. In \textit{International Conference on Learning Representation}.

\end{document}